\pdfoutput=1

\documentclass[11pt]{article}

\usepackage[preprint]{acl_latex}

\usepackage{times}
\usepackage{latexsym}

\usepackage[T1]{fontenc}

\usepackage[utf8]{inputenc}

\usepackage{microtype}

\usepackage{inconsolata}

\usepackage{graphicx}

\usepackage{amsmath}

 \usepackage{booktabs}

\title{
Biases Propagate in Encoder-based Vision-Language Models: A Systematic Analysis From Intrinsic Measures to Zero-shot Retrieval Outcomes
}

 \author{Kshitish Ghate \\
   Carnegie Mellon University \\
  \texttt{kghate@andrew.cmu.edu} \\\And
   Tessa Charlesworth \\
   Northwestern University\\
\texttt{tessa.charlesworth@kellogg.northwestern.edu} \\\AND
  Mona Diab\\
   Carnegie Mellon University\\
  \texttt{mdiab@andrew.cmu.edu} \\\And
  Aylin Caliskan\\
   University of Washington\\
  \texttt{aylin@uw.edu}
}

\usepackage{algorithm}
\usepackage{algorithmic}

\begin{document}
\maketitle
\begin{abstract}
To build fair AI systems we need to understand how social-group biases intrinsic to foundational encoder-based vision-language models (VLMs) manifest in biases in downstream tasks. In this study, we demonstrate that intrinsic biases in VLM representations systematically ``carry over'' or propagate into zero-shot retrieval tasks, revealing how deeply rooted biases shape a model’s outputs. We introduce a controlled framework to measure this propagation by correlating (a) intrinsic measures of bias in the representational space with (b) extrinsic measures of bias in zero-shot text-to-image (TTI) and image-to-text (ITT) retrieval. Results show substantial correlations between intrinsic and extrinsic bias, with an average $\rho$ = 0.83 $\pm$ 0.10. This pattern is consistent across 114 analyses, both retrieval directions, six social groups, and three distinct VLMs. Notably, we find that larger/better-performing models exhibit greater bias propagation, a finding that raises concerns given the trend towards increasingly complex AI models. Our framework introduces baseline evaluation tasks to measure the propagation of group and valence signals. Investigations reveal that underrepresented groups experience less robust propagation, further skewing their model-related outcomes.
\end{abstract}

\section{Introduction}

Modern encoder-based vision-language models (VLMs) such as CLIP \cite{radford2021learning} and BLIP-2 \cite{li2023blip} excel at mapping images and text into a shared representational space. These models are the backbone of many zero-shot applications, from object recognition to image retrieval \cite{Rombach_2022_CVPR, Briggs_Laura_2022, taesiri2022clip, bui2023zero, barraco2022unreasonable,pirom2022object}. However, it is increasingly clear that these representations may contain intrinsic biases  \cite{caliskan2017semantics, charlesworth2022word, caliskan2022gender} for instance, systematically associating certain social groups with negative concepts. The potential for these biases to manifest in downstream applications can undermine fair and ethical treatment across social groups \cite{ghosh-caliskan-2023-person, wolfe2022american}.  There exists a significant gap in understanding how biases intrinsic to the representational spaces of foundational encoder-based VLMs like CLIP \cite{radford2021learning} and BLIP2 \cite{li2023blip} propagate to tasks they perform effortlessly without retraining or fine-tuning. Concretely: the current research quantifies the propagation of biases in encoder-based VLMs by comparing (a) intrinsic bias in the representational space learned by these models to (b) the outcomes of downstream zero-shot tasks, specifically text-to-image (TTI) and image-to-text (ITT) retrieval. 

In our study, the term ``propagation'' refers to the directional flow of biases from model representations to outcomes in downstream tasks. By employing a rigorously controlled experimental design, we minimize external confounding factors, enabling a clear observation of how biases are transferred within encoder-based VLMs. While we recognise this does not fully establish causality, our findings provide robust evidence of systemic bias transfer, justifying the use of the term ``propagation.''

\begin{figure*}[!ht]
\centering
\includegraphics[width=1\linewidth]{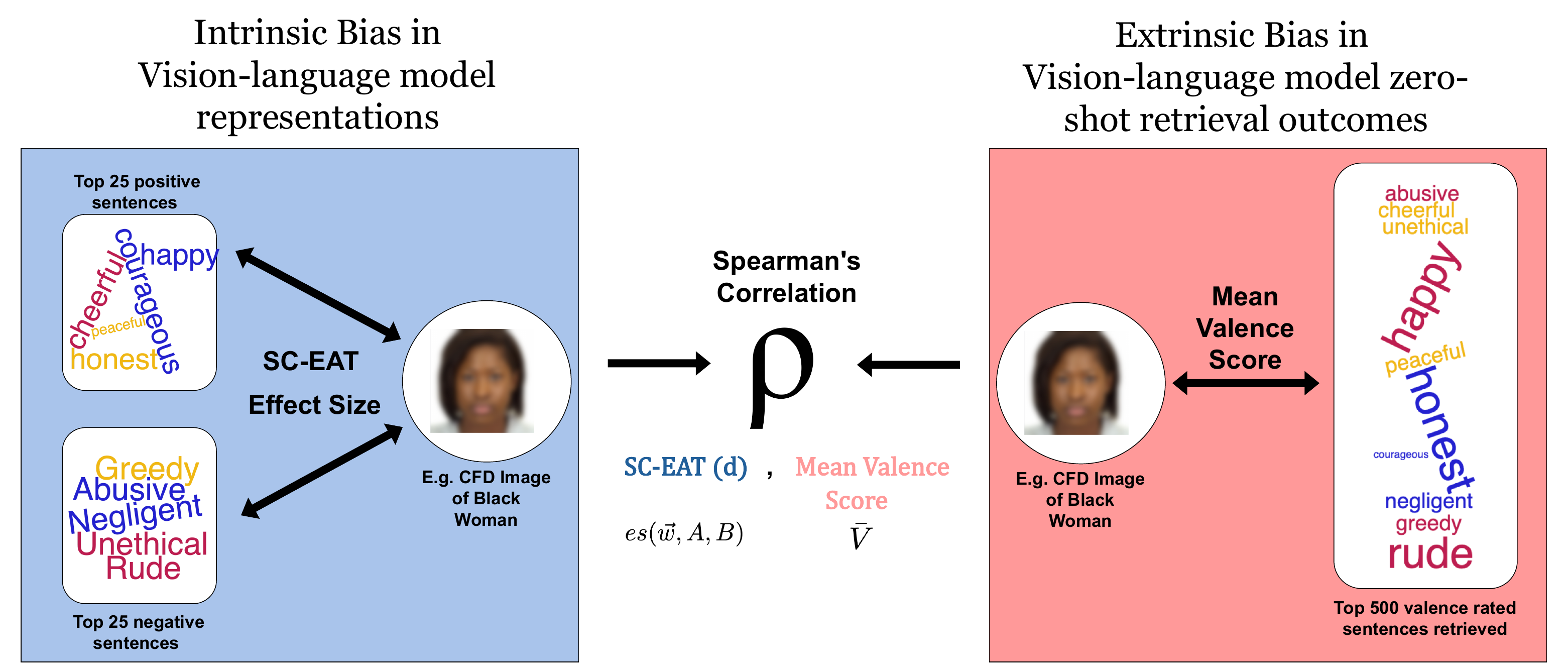}

  \caption{The setup to analyze valence-based bias propagation in Image-to-Text retrieval. Propagation is measured by the correlation between the intrinsic bias (measured by valence SC-EAT), and extrinsic bias (measured by mean valence of retrieved text) for images belonging to a social group. The intrinsic SC-EAT effect size is computed based on the differential association of each CFD image with the top 25 positive and top 25 negative words in sentence templates. The extrinsic metric is derived from the mean valence of the top 500 words in sentence templates retrieved by each image. Ground truth valence scores come from human rated sources \cite{mohammad2018obtaining}. Spearman’s $\rho$ is used to compute the correlation between intrinsic and extrinsic metrics.
  }
   \label{fig:1}
\end{figure*}

To conduct our analysis, we curate balanced datasets comprising images from the Chicago Face Database (CFD) \cite{ma2015chicago} and the Open Affective Standardized Image Set (OASIS) \cite{kurdi2017introducing}, capturing social group signals and valence signals, respectively. Valence is a primary dimension in our analysis, as it captures biased attitudes critical to understanding bias manifestation in AI \cite{toney-caliskan-2021-valnorm, osgood1957measurement}. In parallel, we develop experimental text datasets using semantically-neutral sentence templates \cite{tan2019assessing}, incorporating target words representing either valence signals from the NRC-VAD lexicon \cite{mohammad2018obtaining} or social group signals from prior research \cite{charlesworth2022historical, charlesworth2024extracting}. We measure bias propagation by quantifying both intrinsic and extrinsic biases in encoder-based VLMs. Intrinsic bias is assessed using the single category-embedding association test (SC-EAT) \cite{caliskan2017semantics, caliskan2023artificial}, which evaluates associations between social groups and attribute association signals within the model's representational space. Extrinsic bias is evaluated through downstream zero-shot tasks, such as ITT and TTI retrieval, to assess how these intrinsic biases manifest in practical applications. Figure \ref{fig:1} demonstrates the setup to measure valence-based bias propagation in the ITT setting. 

Our study uniquely examines the propagation of both social group and valence biases in encoder-based VLMs within controlled zero-shot settings, addressing the complexity of bias in multimodal contexts across multiple social groups. It may seem intuitive that intrinsic biases in a model’s representations would naturally emerge in its downstream tasks, simply because they share the same embedding space. However, as we will demonstrate, this relationship is not trivial. Different social groups (e.g., “Black Women” vs. “White Men”) can yield significantly divergent correlations. These systematic differences reflect how complex social cues such as ``data marking'' and group underrepresentation \cite{newton2023hypervisibility, wolfe2022markedness} can affect the flow of biases. Accordingly, the present work also contributes empirical clarity to how specific group and valence signals shape the degree of bias propagation in encoder-based VLMs.

Our study makes the following contributions.\footnote{The code and data is available at \url{https://github.com/kshitishghate/bias_prop}}\\
 \textbf{(1)} We provide a controlled experimental framework for systematically quantifying bias propagation. The framework reinforces previous work showing biases in the representational spaces of encoder-based VLMs. It also, for the first time, shows biases in the downstream zero-shot tasks of TTI and ITT retrieval as they directly relate to intrinsic biases. Moreover, the framework and methods shown here are unique in enabling the study of bias propagation at scale - showing consistent conclusions across 114 scenarios, incorporating 6 social groups, 3 models, and 8 experiments, we show correlations are consistently high and significant (Spearman’s $\rho$ of 0.83 $\pm$ 0.10).\\
 \textbf{(2)} We introduce baseline intrinsic evaluation experiments that evaluate the direct propagation of valence and group signals in zero-shot retrieval tasks. These baselines indicate the extent of accurate retrieval and can serve as benchmarks against which to compare the results of the key experiments examining bias in valence-to-group and group-to-valence propagation. Results show a strong baseline correlation for valence propagation, with a Spearman’s $\rho$ of 0.86 $\pm$ 0.04, and for group information propagation, with a $\rho$  of 0.85 $\pm$ 0.12.  \\
\textbf{(3)} The controlled framework will support future empirical and theoretical understanding of AI bias dynamics. Specifically, the results highlight variation in bias propagation across social groups. Marginalised groups that correspond to being underrepresented in data experience less robust propagation and more skewed model-related outcomes. We also show that models have a tendency to show higher bias propagation as they scale in size and performance --- a finding with critical implications during an era of rapid AI expansion.

\section{Related Work}
\label{sec:rw}

\textbf{Bias in Vision-Language Models.}
Recent research on bias in VLMs moves beyond standalone language or vision models to investigate when these two modalities intersect. For instance, \citet{srinivasan-bisk-2022-worst} extended text-based gender bias studies (e.g., \cite{su2019vl}) to the multi-modal setting. Using VL-BERT, they demonstrate that VLMS tend to reinforce and exaggerate gender biases and stereotypes. \citet{wolfe2022american} evaluated biases associating American with White, as seen in foundational cognitive science studies \cite{devos2005american} in CLIP using embedding association tests \cite{caliskan2017semantics} with the CFD \cite{ma2015chicago}. They observed that pictures of White faces (versus Asian, Latina/o, and Black faces) had a higher association with collective in-group words. \citet{zhou2022vlstereoset} assessed multiple VLMs with VLStereoSet, and found pervasive gender and race stereotypes and biases, which are more complex in VLMs than in pre-trained language models. \citet{wolfe2023contrastive} focused on how VLMs, trained on web-scraped data, often perpetuate the sexual objectification of women and girls. 

A related strand of work by investigates the origins of bias in VLMs by examining the influence of pretraining factors. \citet{berg-etal-2022-prompt} compare gender bias across 9 CLIP models and find that larger pretraining datasets often associated with better down-stream performance, also show the least bias. \citet{ghate-etal-2025-intrinsic} extend this study across 131 CLIP models, more than 20 architectures and pretraining datasets and find that the choice of the pretraining dataset and data filtering strategy used to filter it is the most significant predictor of intrinsic bias in CLIP-based VLMs, over and above variables such as architecture choice, size of the pretraining dataset and number of model parameters. \citet{hong2024s} audit CLIP-based pretraining data filtering strategies and find that data related marginalized groups are subject to higher rates of exclusion compared to historically well-represented Western-centric demographics.

\noindent \textbf{Bias Propagation in Pretrained models.} Several studies have investigated how biases intrinsically present in pretrained language  models can influence downstream model outcomes. \citet{cao-etal-2022-intrinsic} highlighted a weak correlation between intrinsic and extrinsic fairness metrics for contextualized language models in supervised settings. \citet{steed2022upstream} investigated the bias transfer hypothesis, which posits that biases in large language models (LLMs) transfer into harmful task-specific behavior after fine-tuning. Their findings indicate that reducing intrinsic bias before fine-tuning does little to mitigate discriminatory outcomes. However, it is crucial to note that the downstream tasks examined in such past work are not zero-shot. Instead, these tasks involve fine-tuning on specific datasets which can lead the model to learn spurious associations inherent in those datasets. \citet{feng-etal-2023-pretraining} explored the propagation of political biases in language models. Their work revealed that pre-trained language models possess political leanings, influenced by politics in training corpora, which affects outcomes in downstream tasks. 
\citet{cabello-etal-2023-evaluating} showed intrinsic biases do not consistently correlate with extrinsic biases in fine-tuned tasks of encoder-based VLMs. They proposed gender-neutral pretraining as a mitigation strategy and demonstrated its ability to reduce group disparities. However, their work also focused on finetuning outcomes and lacked a systematic framework to directly investigate bias propagation.

\noindent In contrast, we focus on zero-shot tasks, where models are directly examined for bias propagation from intrinsic to extrinsic levels without additional fine-tuning, and models directly rely on their pretrained representations. We introduce a controlled framework to investigate biases across multiple VLMs and social groups. This enables us to uncover previously overlooked systematic trends in bias propagation.
\section{Data}
\label{sec:data}\begin{table*}[!ht]
\centering
\caption{Summary Dataset Statistics and Usage in Intrinsic and Extrinsic Measures. Further details in Appendix \ref{sec:appData}.}
\footnotesize
\label{tab:dataset-statistics}
\resizebox{\textwidth}{!}{%
\begin{tabular}{lccc}
\toprule
\hline
\textbf{Dataset} & \textbf{Size} & \textbf{Intrinsic Measure} & \textbf{Extrinsic Measure} \\ \hline
\textbf{CFD} & 597 images & Group-based images SC-EAT & Group representation in retrieved images \\ 
\textbf{OASIS} & 900 images & Valenced images in SC-EAT  & Valence ratings in retrieved images \\ 
\textbf{NRC-VAD Lexicon} & 20,000 words & Valenced Text SC-EAT & Valence ratings of retrieved text \\ 
\textbf{Group Labels} & 864 phrases  & Group-based text in SC-EAT & Group representation in retrieved text \\ \hline
\bottomrule
\end{tabular}%
}
\end{table*}

We employ carefully curated datasets that provide experimental control and ground truth values.

\noindent \textbf{Chicago Face Database (CFD)  \cite{ma2015chicago}}: The CFD dataset consists of 597 self-identified images of men and women participants belonging to 4 different racial demographics, namely, Black, White, Latino/a and Asian. 
This dataset was selected for its reliable, self-identified human face images with clear group signals, unlike FairFace \cite{karkkainen2021fairface}, which lacks self-identified labels and includes visually noisy images, potentially biasing annotations. We first expand the dataset to facilitate the scale and generalization of our experiments by creating within-group image morphs following the approach in \cite{wolfe2022evidence} as described in Appendix \ref{app:morph}. We then randomly sampled images without replacement.

\noindent \textbf{Open Affective Standardized Image Set (OASIS) \cite{kurdi2017introducing} }: Provides 900 human valence-rated images, enabling the study of how VLMs handle non-group-related valence signals from naturalistic image inputs. The images depict a broad spectrum of themes such as animals, humans, objects, and scenes.

\noindent \textbf{Textual Templates and Lexica}: Our research requires controlled text datasets. Semantically neutral sentence templates enable understanding how the valence and social group properties of target lexica embedded in them influence biases in VLMs. As such, we employ lexica embedded in 6 such templates taken from \cite{may-etal-2019-measuring} for controlled analysis of how words/phrases, particularly those with valence or social group signals, influence biases in extrinsic VLM outputs. For experiments using valence-rated sentences, we employ the NRC-VAD lexicon \cite{mohammad2018obtaining}, comprising approximately 20,000 English words. An example sentence created from the valence lexicon ``happy'' is, ``This is the word happy.'' For experiments with group signals, we employ a curated list of 864 group-label phrases derived from \citet{charlesworth2022historical}. An example sentence created from the group-label ``Black Woman'' is ``This is the word Black Woman.'' We focus on 6 intersectional race and gender groups: ``Asian Men,'' ``Asian Women,'' ``Black Men,'' ``Black Women,'' ``White Men,'' and ``White Women'' in our analyses.

\noindent Summary statistics of CFD, OASIS, NRC-VAD lexicon, and group-labels are present in Table \ref{tab:dataset-statistics}. For their description and usage details in our experiments, please refer to the Appendix \ref{sec:appData}.

\section{Approach}

Social group biases in this study are defined as the association between group signals (e.g., the representation of Black women) and valence signals (e.g., the representation of positivity vs. negativity). This approach is motivated by established methods in the field of AI ethics, such as the Embedding Association Test (EAT) applied to embedding spaces \cite{caliskan2017semantics}, providing a scientifically robust framework for studying biases. Valence is also integral to our analysis as it reflects the positive or negative connotations associated with group representations, and is the basic dimension along which biases are evaluated \cite{eagly1998attitude}.

\subsection{Intrinsic and Extrinsic Bias Measurement}
\textbf{Intrinsic bias} is quantified using the Single Category-Embedding Association Test (SC-EAT) \cite{caliskan2017semantics} effect sizes \cite{cohen2013statistical} which allows for a principled assessment of biases within model representations. SC-EAT quantitatively evaluates the association between a single target stimulus (e.g., an image or word representing a social group) and sets of attribute stimuli (e.g., words or images with valence or group content). The association is measured by the mean cosine similarity between the embeddings of the target and attribute stimuli, normalized by the standard deviation of these similarities across all stimuli. This is quantified as the SC-EAT Cohen’s d effect size given by:

\[
\scriptsize es(\vec{w}, A, B) = \frac{\textrm{mean}_{a \in A} \textrm{cos}(\vec{w}, \vec{a}) - \textrm{mean}_{b \in B} \textrm{cos}(\vec{w}, \vec{b})}{\textrm{std-dev}_{x \in A \cup B} \textrm{cos}(\vec{w}, \vec{x})}
\]

\(\vec{w}\) represents the embedding of the target stimulus, which in our case can represent either a sentence or image of interest, and \(A\) and \(B\) are sets of attributes determined either by valence or group content.

\noindent \textbf{Extrinsic bias} in ITT and TTI is evaluated through two measures derived from work quantifying bias in task outcomes related to valence and content \cite{charlesworth2024extracting, kong2024mitigating}  :

\noindent 1. \textbf{Mean of Valenced Text/Images Retrieved:} This measure calculates the average valence of retrieved text/image by the model for a given prompt:
\[
\bar{V} = \frac{1}{N} \sum_{i=1}^{N} V_{i}
\]

\noindent Where \( \bar{V} \) is the average valence, \( V_{i} \) is the valence of the \( i \)-th retrieved item, and \( N \) is the total number of retrieved items. The measure quantifies the degree to which the model's outputs, in terms of textual descriptions or image retrieval, skew towards positive or negative valence. This measure helps understand how the model's internal biases translate into biases in the valence content it retrieves.

\noindent 2. \textbf{Proportion of Text/Images Belonging to a Social Group Retrieved:} This measure evaluates the likelihood or proportion of the model retrieving text or images associated with specific social groups in response to given prompts, calculated as:

\[
P_{grp} = \frac{\textrm{Number of group-related items retrieved}}{\textrm{Total number of retrieved items}}
\]
 
\noindent where $P_{grp}$ is the proportion expected to remain constant across groups for a prompt. An unequal $P_{grp}$ when the prompt is expected to be associated with all groups equally shows model bias.

\subsection{Framework to Measure Bias Propagation}

To evaluate how intrinsic biases in VLMs propagate to downstream retrieval tasks, we propose a unified framework (Algorithm \ref{alg:unified_framework}) where the following variables are defined:

\noindent \textbf{Data ($D$):} The input can be any of our curated datasets (CFD/OASIS images or text templates using NRC-VAD/group labels).

\noindent\noindent\textbf{Retrieval Direction ($R$):} Specifies the zero-shot retrieval task, either ITT or TTI.

\noindent \textbf{Content Type ($C$):} Declares whether the current bias analysis targets valence (positivity/negativity) or group (e.g., race and gender).

\noindent\textbf{Correlation ($\rho$):} The final Spearman’s correlation computed over intrinsic and extrinsic metrics.

\begin{algorithm*}[tb]
\caption{Unified Bias Propagation Framework}
\label{alg:unified_framework}
\begin{algorithmic}[1]
\STATE \textbf{Input:} Data $D$, Retrieval Direction $R$, Content Type $C$
\STATE \textbf{Output:} Correlation $\rho$ between intrinsic and extrinsic metrics
\STATE $intrinsic\_metric \gets$ \textbf{CalculateIntrinsicAssociations}($D$, $C$)
\IF{$R == \text{image\_to\_text}$}
    \STATE $retrieved\_items \gets$ \textbf{RetrieveTextFromImage}($D$)
\ELSIF{$R == \text{text\_to\_image}$}
    \STATE $retrieved\_items \gets$ \textbf{RetrieveImagesFromText}($D$)
\ENDIF
\STATE $extrinsic\_metric \gets$ \textbf{CalculateExtrinsicOutput}($retrieved\_items$, $C$)
\STATE $\rho \gets$ \textbf{SpearmanCorrelation}($intrinsic\_metric$, $extrinsic\_metric$)
\STATE \textbf{return} $\rho$
\end{algorithmic}
\end{algorithm*}

We now provide a step-by-step overview the functions that constitute our framework in \ref{alg:unified_framework} followed by an example.

\noindent \textbf{1. CalculateIntrinsicAssociations}($D$, $C$): This function computes the intrinsic association metric based on the content type $C$ and data $D$. If $C$ is ``valence,'' the metric is calculated using the SC-EAT on valence signals (e.g., from the OASIS dataset for images or the NRC-VAD lexicon for text). If $C$ is ``group,'' the SC-EAT is applied to group signals (e.g., from the CFD for images or predefined group labels for text).

\noindent \textbf{2. RetrieveTextFromImage}($D$): Retrieves top $k$ sentences associated with input $D$ using a VLM.

\noindent \textbf{3. RetrieveImagesFromText}($D$): Retrieves top $k$ images associated with input $D$ using a VLM. 

\noindent \textbf{4. CalculateExtrinsicOutput}($retrieved\_items$, $C$): This function calculates the extrinsic bias metric based on the retrieved items and the content type $C$. If $C$ is ``valence,'' the metric is the mean valence of the retrieved items. If $C$ is ``group,'' the metric is the proportion of retrieved group items.

\noindent \textbf{5. SpearmanCorrelation}($intrinsic\_metric$, $extrinsic\_metric$): This function calculates the Spearman's correlation $\rho$ between the intrinsic and extrinsic metrics, reflecting the degree of bias propagation. We choose to employ Spearman's $\rho$ due to its non-parametric nature and ability to capture nonlinear relationships \cite{spearman1961proof}.

\noindent\textbf{Mini-Example:} Suppose we want to see whether valence bias toward ``Black Women'' is reflected in ITT retrieval (depicted in Figure \ref{fig:1}). Here, input $D$: 6000 images of (1000 per social group) from CFD; 20,000 words in sentence templates from NRC-VAD. C: ``Valence''; R: ITT. Concretely:

\noindent \textbf{1. Intrinsic Association:} We apply SC-EAT to 1000 images of Black Women from CFD. Each image’s embedding is compared with the top-25 positive vs. negative words in sentence templates (from the NRC-VAD). The result is an intrinsic valence effect size per image.

\noindent \textbf{2. ITT Retrieval:} We then feed each of those 1000 images into our VLM and retrieve the top-$500$ text items (e.g., ``This is the word happy,'' ``This is the word sad,'' etc.).

\noindent \textbf{3. Extrinsic Metric:} For each image, we measure the average valence of the retrieved text (are the words mostly positive or negative?).

\noindent\textbf{4. Correlation:} We measure the correlation between the valence effect sizes and the mean retrieved valence over the 1000 images. If images with high negative SC-EAT scores also retrieve mostly negative text descriptors, it suggests strong bias propagation (likely a high $\rho$).

\subsection{Measuring Bias Propagation Via Valence and Group Signals}
\label{prop_approach}

The core of our experimental design to analyze bias propagation consists of 8 experiments that all follow the framework in Algorithm \ref{alg:unified_framework}, and are evenly divided to differ in: (1) direction (i.e., 4 experiments test ITT association and bias propagation, while 4 test TTI association and bias propagation); and (2) valence versus group content (i.e., 4 experiments test the propagation of valence, while 4 test the propagation of group signals). The application of Algorithm~\ref{alg:unified_framework} is explained in the following propagation experiments. For all experiments, the correlation is computed over intrinsic and extrinsic metric scores for target group of images or text per model for ITT and TTI respectively. Exact details can be found in Appendix Figure \ref{fig:doe}.

\noindent  \textbf{Baseline Valence-Valence Signal Propagation (1*-a and 1*-b)}: This is a baseline experiment to measure how intrinsic valence signals propagate to valence outcomes. The framework parameters are --  $D$: OASIS images and NRC-VAD lexicon sentences; $C$: ``valence.'' We label the $R$: ITT direction as (1*-a) and the $R$: TTI direction as (1*-b). Intrinsic metrics: SC-EAT associations for top valenced images/text. Extrinsic metrics: Mean valence ratings of top-500 retrieved images/text. For all experiments, we choose a retrieval $k$ of 500 items as it provides a significant sample size that is diverse while being computationally feasible.

\noindent  \textbf{Baseline Group-Group Signal Propagation (2*-a and 2*-b)}: This is a baseline experiment to measure how intrinsic group signals propagate to group outcomes using CFD images and predefined group labels. The framework parameters are --  $D$: CFD images and group label sentences; $C$: ``group.'' We label the $R$: ITT direction as (2*-a) and the $R$: TTI direction as (2*-b). Intrinsic metrics: SC-EAT group associations. Extrinsic metrics: Proportion of correctly identified group representations in the top-500 retrieved images/text. 

\noindent  \textbf{Valence-to-Group Signal Propagation (1-a and 1-b)}: Assesses how valence signals influence group-related retrieval outcomes. The framework parameters are -- We label the $R$: ITT direction as (1-a) with the inputs $D$: CFD images and NRC-VAD lexicon sentences; $C$: ``valence.'' The $R$: TTI direction is labeled as (1-b) with the inputs $D$: group label sentences and OASIS images; $C$: ``valence.'' Intrinsic metrics: SC-EAT valence effect sizes for group-associated images/text. Extrinsic metrics: Mean valence or proportion of group-identified content in the top-500 retrieved images/text. Figure \ref{fig:1} outlines the ITT version of these experiments visually.  

\noindent  \textbf{Group-to-Valence Signal Propagation (2-a and 2-b)}: Investigates the influence of group signals on valence-related retrieval outcomes. The framework parameters are -- We label the $R$: ITT direction as (2-a) with the inputs $D$: OASIS images and group label sentences; $C$: ``group.'' The $R$: TTI direction is labelled as (2-b) with the inputs $D$: NRC-VAD lexicon sentences and CFD images; $C$: ``group.''  Intrinsic metrics: SC-EAT group associations for valenced images/text. Extrinsic metrics: Mean valence or proportion of group-related content in the top-500 retrieved images/text.

\section{Experiments and Results}
\label{experiments}

\subsection{Evaluation of Intrinsic Biases in VLMs} 
\label{Initialexp}

To lay the groundwork for our study, we initially assess intrinsic biases present in the representational spaces of encoder-based VLMs using valence and group-based SC-EATs. Details of the experimental setup for measuring these biases in experiments 1-a, 1-b (see Figure \ref{fig:1} for a visualization of the setup for ITT), 2-a, and 2-b will be elaborated later in this section, ensuring a clear linkage between initial findings and their implications in broader model behaviors. We choose to evaluate encoder-based VLMs, specifically, OpenAI versions of CLIP-B-32, CLIP-L-14, and Salesforce's BLIP-2 (the encoder-decoder versions) detail the choice of models used in our experiments due to their foundational role and extensive usage in the community (over 50 million downloads on Huggingface \footnote{https://huggingface.co/openai}), and availability of principled methods \citet{steed2021image} to measure biases within their representations. Further model details are elaborated in the Appendix \ref{sec:Models}.  

Our analysis highlights significant intrinsic biases within VLMs. Specifically, the aggregate valence-based SC-EAT effect sizes for different social groups obtained in experiments 1-a and 1-b reveal that ``Black Women,'' followed by ``Black Men'' are associated with the most negative valence (with -1.43 and -1.31 z-scored effect sizes (d) respectively). ``White Men'' and ``White Women'' are similarly placed with 0.44 and 0.45 z-scored effect sizes respectively, while ``Asian Women'' are most associated with positive valence with 1.21 z-scored effect size, indicating a strong valence-based representation bias. 2-a and 2-b demonstrate an overrepresentation of ``White Men''  and ``White Women'' group associations to valenced text and images (with 0.48 and 0.46 z-scored effect sizes respectively) as opposed to ``Black Women'' (with -0.58 z-scored effect size). The detailed results, including figures highlighting these findings are presented in the Appendix \ref{app:initial_res}.

\begin{figure}[ht]
  \centering
\centering\includegraphics[width=0.8\linewidth]{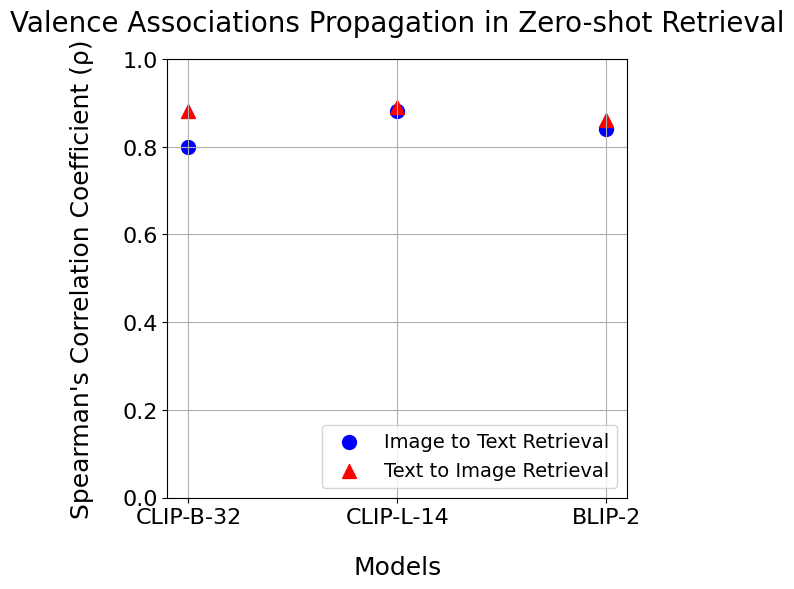}

\centering
  \caption{Exp. 1*-a and 1*-b illustrate Spearman's $\rho$ measuring the correlation of intrinsic valence associations with extrinsic valence outcomes in zero-shot retrieval tasks. }
  \label{fig:exp-1*}
\end{figure}

\begin{figure*}[ht!]
      \begin{minipage}[c]{0.46\textwidth}
\centering
\includegraphics[width=0.9\linewidth]{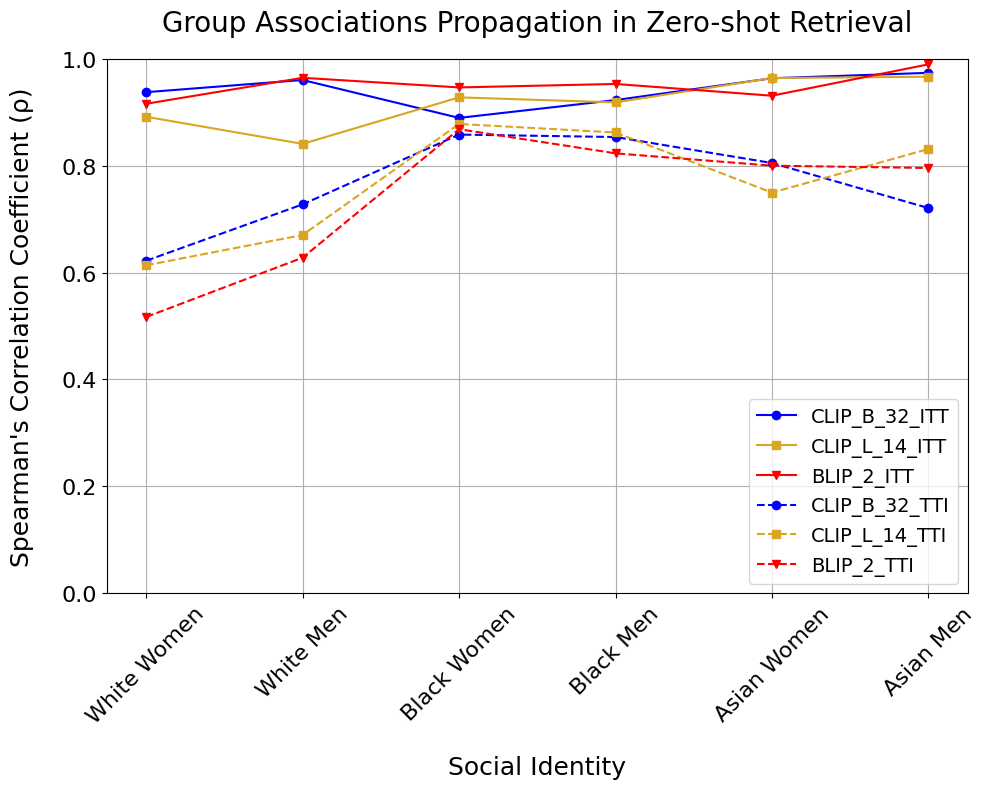}
\centering
  \caption{Exp. 2*-a and 2*-b illustrate Spearman's $\rho$ measuring the correlation of intrinsic group signal propagation to extrinsic group outcomes in zero-shot retrieval tasks, stratified by social groups. }
  \label{fig:exp-2*}
       \end{minipage}
\begin{minipage}[c]{0.49\textwidth}
\centering  

\includegraphics[width=0.9\linewidth]{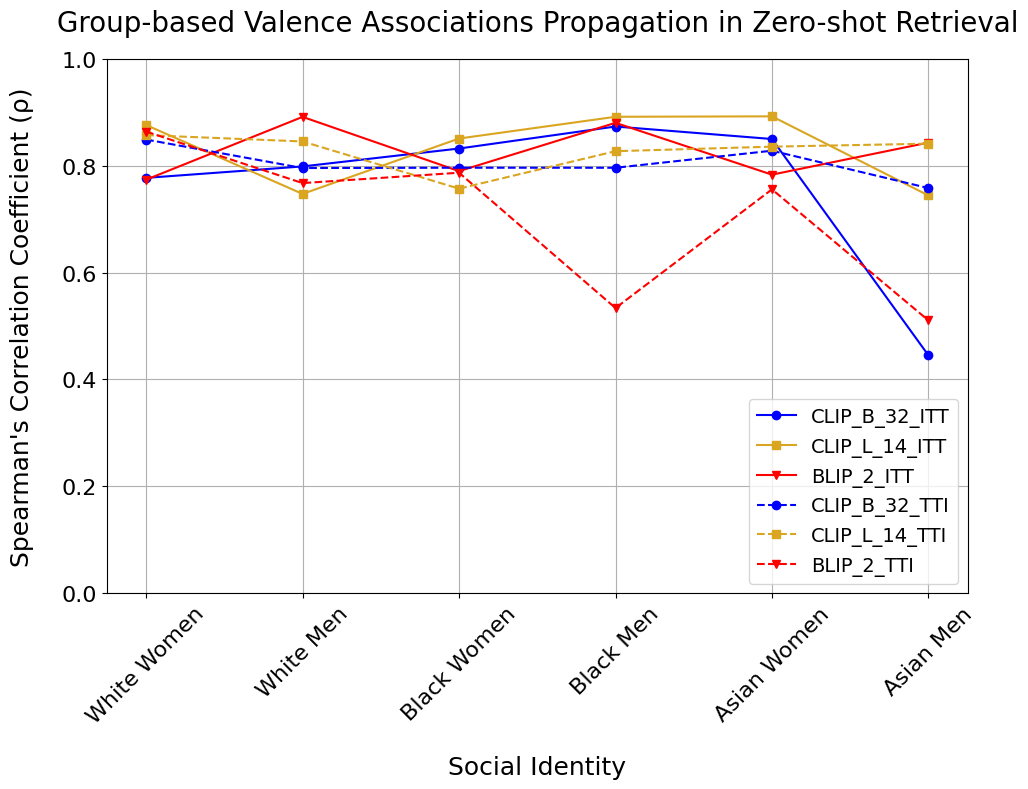}
  \caption{Exp. 1-a and 1-b illustrate Spearman's $\rho$ measuring the correlation of intrinsic valence-based bias propagation to extrinsic valence outcomes in zero-shot retrieval tasks, stratified by social groups. }
  \label{fig:exp-1}

  \end{minipage}
  \hspace{8mm}
\end{figure*}

\begin{figure}[ht]
  \centering
\centering

\includegraphics[width=0.9\linewidth]{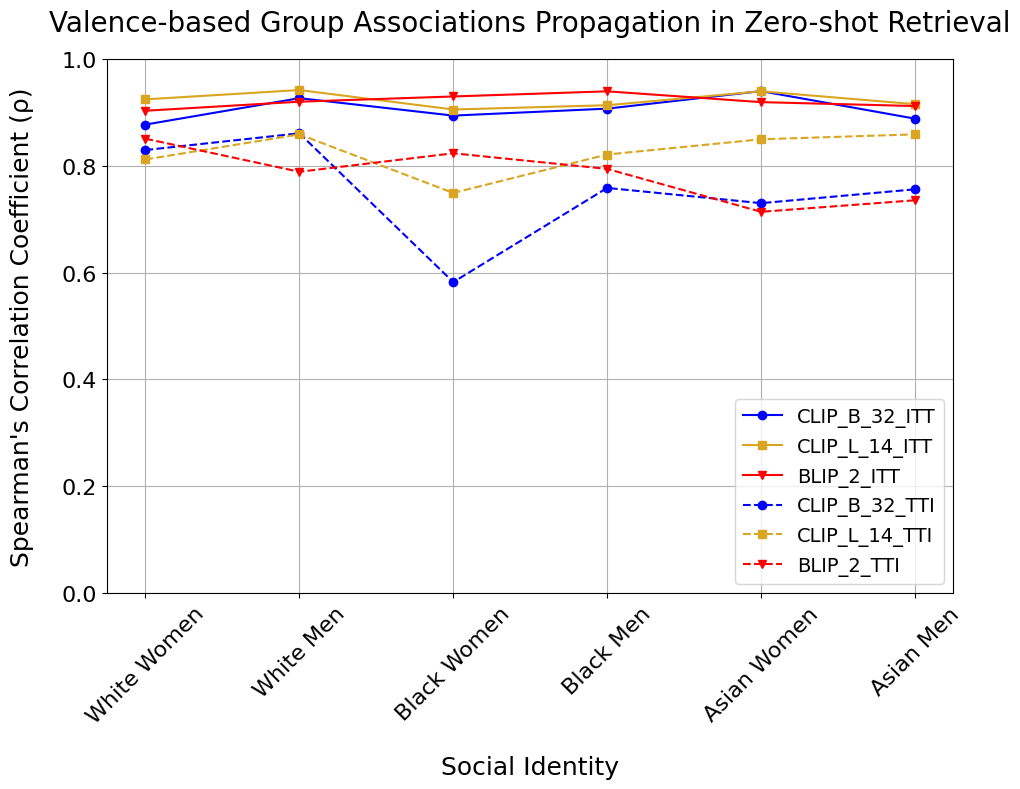}

  \caption{Exp. 2-a and 2-b illustrate Spearman's $\rho$ measuring the correlation of intrinsic valence-based group bias propagation to extrinsic group outcomes in zero-shot retrieval tasks, stratified by social groups.}
  \label{fig:exp-2}
\end{figure}

\subsection{Bias Propagation Via Valence and Group Signals}
We now list the key findings from our propagation experiments of Section \ref{prop_approach}. Appendix \ref{sec:appExp} has further findings and reproducibility details.

\noindent  \textbf{Baseline Valence-Valence Signal Propagation (1*-a and 1*-b)}: In 1*-a and 1*-b (Figure \ref{fig:exp-1*}), the models show a strong aggregate Spearman's correlation of 0.84 $\pm$ 0.04 $\rho$ in 1*-a,  0.87 $\pm$ 0.02 $\rho$ in 1*-b.  Furthermore, we observed larger models like CLIP-L-14 consistently showing higher correlations (up to $\rho = 0.88$) compared to smaller models like CLIP-B-32 ($\rho = 0.80$).

\noindent  \textbf{Baseline Group-Group Signal Propagation (2*-a and 2*-b)}:  2*-a and 2*-b demonstrate that VLMs can robustly propagate social group signals, as indicated by the strong aggregate Spearman's correlations (0.94 $\pm$ 0.04 $\rho$ in 2*-a,  0.76 $\pm$ 0.11 $\rho$ in 2*-b) observed across all models (Figure \ref{fig:exp-2*}).

\noindent  \textbf{Valence-to-Group Signal Propagation (1-a and 1-b)}: 1-a and 1-b results in Figure \ref{fig:exp-1} demonstrate significant positive aggregate Spearman's correlations across models and social groups with 0.81 $\pm$ 0.10 $\rho$ in 1-a, and  0.78 $\pm$ 0.10 $\rho$ in 1-b.

\noindent  \textbf{Group-to-Valence Signal Propagation (2-a and 2-b)}: 2-a and 2-b results in Figure \ref{fig:exp-2} demonstrate significant positive aggregate Spearman's correlations across models and social groups with 0.91 $\pm$ 0.02 $\rho$ in 2-a, and 0.78 $\pm$ 0.07 $\rho$ in 2-b.

\section{Discussion}

Our study first quantifies the intrinsic biases in encoder-based VLMs. Next, across 8 experiments and 114\footnote{We clarify that we have 114 analyses due to no group-based distinctions in our valence-valence propagation baseline (2-scenarios * 3-models  + 6-scenarios * 3-models * 6-social groups).} analyses, the current work finds consistently across all approaches, positive and generally strong correlations (Spearman's $\rho$ of 0.83 $\pm$ 0.10) between intrinsic biases in VLM representational spaces and extrinsic bias outputs in zero-shot ITT and TTI retrieval tasks, with notable systematic differences relating to social group identity.  

\noindent \textbf{Evaluation of Intrinsic Biases in VLMs.} Our experiment on initially quantifying intrinsic bias demonstrates the intersectional gender hypothesis \cite{ghavami2013intersectional} where biases between men and women are most similar in the case of biases between ``White Men'' and ``White Women.'' Additionally, aggregate group SC-EAT effect sizes across experiments 2-a and 2-b indicate ``White Men'' and ``White Women'' have the highest levels of group associations to valenced content in VLMs, meaning they are overrepresented in model outcomes. In contrast, ``Black Women'' demonstrate minimal intrinsic group associations in 2-a and 2-b while being associated with the most negatively valenced content in 1-a and 1-b. This is consistent with research on group identity bias \cite{devos2005american,wolfe2022american}

\noindent \textbf{Baseline Propagation of Valence and Group Signals.} Experiments 1*-a and 1*-b show that VLMs robustly encode and propagate fundamental valence signals indicating that valence is a significant signal in vision and language domains \cite{wolfe2022vast, toney-caliskan-2021-valnorm}. However, VLMs show distinct biases in how they propagate different social group representations, for instance, the ``Black Women'' group showed a particularly consistently high correlation (e.g. $\rho = 0.86$ in 2*-b for BLIP-2), while the ``White Women'' group showed a lower correlation (e.g. $\rho = 0.53$ in 2*-b for BLIP-2). This result could be attributed to the ``marking'' of certain identities in datasets, especially to non-default (i.e., non-White) groups. Marking and ``othering'' can lead to more distinct overfitted portrayals that increase propagation and risks entrenching stereotypes \cite{newton2023hypervisibility, wolfe2022markedness}.

\noindent To support this result, we found that the bigram ``Black Women'' occurs more frequently in English language compared to ``White Women'' according to Google Ngram (\url{https://books.google.com/ngrams/}) statistics from the last century. The lower correlation for ``White Women'' presents a complex issue. The default category for a person in AI is White \cite{wolfe2022american}. Accordingly, the default for Woman is a White Woman \cite{wolfe2022markedness}. Consequently, references to ``White Women'' typically just include the word women. As a result, our analysis using the input ``White Women'' likely captures a weaker representation compared to ``Black Women.'' as they are not distinctly ``different'' enough to be as marked as groups like ``Black Women,'' yet they are not represented as the ``default'' group like ``White Men,'' causing the models to learn underfitted representations of them. Following existing audits of VLM-based data filtering methods \cite{hong2024s}, we leave the analysis of disentangling further societal defaults in AI bias analysis to future work.

\noindent We also note that propagation in all cases is not 100\% due to variable template content and confounding signals in the text and images, and this is seen more so in TTI as opposed to ITT. The generally high correlations we observe suggest that when models operate purely on their learned representations, particularly without the intervening layer of task-specific fine-tuning that can introduce new associations, the biases embedded in the representational space directly manifest in outputs.

\noindent \textbf{Multidimensional Nature of Bias in Propagation through Intersection of Valence and Group Signals.} VLMs effectively process simple valence signals, their handling of complex social group signals is inconsistent, which would imply skewed representations and degrading performance. For instance, in 1-b, ``White Women'' showed strong correlations in all models, with the highest being in BLIP-2 at 0.86 $\rho$. In contrast, ``Asian Men'' and ``Black Men'' exhibited lower correlations in some models, particularly in BLIP-2 (0.51 $\rho$ and 0.53 $\rho$, respectively). These low correlations were weaker than ``White Men'' and also ``Asian Women'' or ``Black Women,'' respectively, suggesting the intersectional combination of men with a marginalized racial identity may reduce Valence-to-Group propagation in TTI.  This may be attributed to the underrepresentation of ``Asian Men'' and ``Black Men'' in the selected models. Google Ngram statistics suggest that mentions of these terms are far less common than groups such as ``White Men.'' 
 
\noindent Figures \ref{fig:exp-1} and \ref{fig:exp-2} highlight the subtle yet important differences in how biases observed in the initial intrinsic bias assessment propagate to zero-shot retrieval tasks. Notably, ``Black Women,'' who exhibit the least magnitude in group associations to valenced content in 2-a and 2-b show significantly varied outcomes in Figure \ref{fig:exp-2}'s Group-to-Valence association propagation. This variability can be linked to the inconsistent representations of the complex intersectional interactions of group identity and valence in VLMs. For instance, lower correlations may signal a model's inability to consistently represent content associated with certain social groups and impact downstream performance.

\noindent \textbf{Scaling of Models and Bias Propagation}.
We found that model size and complexity play a role in bias propagation. As seen in 1*-a and 1-a, larger and better-performing models like CLIP-L-14, which regularly outperforms CLIP-B-32 on numerous zero-shot benchmarks \cite{radford2021learning}, demonstrated stronger correlations in bias propagation. Similarly, in the case of ``White Women,'' the CLIP-B-32 model showed a Spearman's correlation of 0.78 $\rho$ in 1-a, while CLIP-L-14 model demonstrated a higher correlation of 0.88 $\rho$ suggesting that as models scale, the degree of valence and group bias propagation increases. While this enhanced propagation capability can be advantageous in terms of the ``accuracy'' of the models for simple Valence-to-Valence or Group-to-Group tasks, it also implies a heightened risk of replicating and amplifying biases in more complex scenarios such as in 1-a. This finding of strong bias propagation in larger models in previous work that demonstrate language models that perform better on standard benchmarks have a greater risk of toxic generations \cite{longpre-etal-2024-pretrainers}, and encoder-based CLIP models exhibit greater intrinsic bias as they scale \cite{ghate-etal-2025-intrinsic}. Targeted bias mitigation strategies are thus essential for larger models through training and reinforcement learning-based approaches where fair intrinsic group representation objectives are prioritized. Future work can also look to develop flexible learning algorithms that will inhibit and steer models to equitable outcomes.

\section{Conclusion}

This research investigated intrinsic biases propagation in ITT and TTI retrieval of encoder-based VLMs. Across 114 analyses focusing on six social groups, we demonstrated that bias propagate with high and systematically varying correlations (Spearman's $\rho$ of 0.83 $\pm$ 0.10 on average). Larger and more complex models showed greater propagation of biases, a finding that is particularly salient given the deployment of increasingly larger AI models. Our results imply detrimental downstream performance implications for marginalised groups that are misrepresented in these models.
\section{Limitations}

The image datasets used, notably the Chicago Face Database (CFD) and OASIS, provide a substantial foundation for analyzing biases as they provide self identification information and human-rated scores. However, they are far from representing the full complexity of human identities and valenced stimuli. In particular, the reliance on these datasets created in the United States could inadvertently reinforce a Western-centric perspective, and fail to account for cultural differences in how facial images and valenced stimuli are experienced. 

In this study, we focus on 3 well-known encoder-based VLMs; CLIP-B-32 and CLIP-L-14 to capture a realistic scaling effect (150M vs. 430M parameters), plus BLIP-2 for architectural diversity. Future work that evaluates more encoder-based VLMs and even newly emerging models is straightforward with our controlled framework, although outside the current scope of the study’s computational limits. Furthermore, our focus on English-only VLMs is an added limitation, as the interpretation of social groups and valence can vary across different languages and cultures \cite{charlesworth2023project}. Such an English-centric perspective could limit understanding of the global impact of AI biases, highlighting a need for multi-lingual and culturally diverse research in future studies. Furthermore, our 

The use of `propagation' in our study denotes the directional flow of biases from model representations to outcomes in downstream tasks. This terminology is supported by the rigorously controlled experimental design that minimizes external confounding factors, allowing a clear observation of how biases are transferred within VLMs. Our comprehensive analysis across 114 scenarios reveals consistent correlations between intrinsic and extrinsic biases, substantiating the directional influence implied by propagation. Pinpointing the exact origin of these biases (e.g., training data vs. architecture) is distinct from measuring if and how biases appear in outputs. Our focus is on the latter which is to systematically demonstrate that biases measured intrinsically do not stay hidden in the intrinsic representations, but are in fact manifested in real zero-shot tasks. Future research may further delineate the precise causal mechanisms along the lines of \citet{ghate-etal-2025-intrinsic}, for instance, the nature of text to image correspondence, pretraining data composition, model objective, and architectural constraints, which are outside the scope of our current study. Using our study’s findings to develop methods to debias and steer model representations to be fairer and validated in further downstream tasks such as text-to-image generations presents another important avenue for exploration.

Our methodology to measure intrinsic associations focuses on SC-EAT because it is a principled and validated embedding-association technique which isolates how one target category associates with two sets of attributes. The method is grounded in cognitive science and social psychology literature \cite{greenwald1998measuring} while other existing methods are ad hoc \cite{blodgett-etal-2020-language} and require further grounding. Additionally, zero-shot retrieval is a primary use-case for these models in practice, which we choose to evaluate using our chosen extrinsic metric. In future work, we hope to see the exploration of ML tasks and downstream settings that may correspond to other bias and fairness metrics (for instance, metrics that exist for fairness in binary classification). However, our choice of extrinsic bias metric here is most aligned with measuring representational harms from zero-shot retrieval tasks.

Finally, our approach to intersectionality, while an attempt to address complex social identities, still only captures a selection of 6 intersectional identities. There are hundreds of potential intersecting identities, including identities related to health, occupation, class, and more \cite{nicolas2023valence}. Future studies could explore more intersections beyond gender and race. It is important to note, however, that our methods to quantify bias propagation are generalizable and can be easily adapted to analyze any group associations.

\section{Ethical Considerations}

As the use of VLMs becomes more widespread in various industries, including object tracking, robot training, advertising, and marketing, \cite{Briggs_Laura_2022, barraco2022unreasonable, bui2023zero, taesiri2022clip} it is important to consider the ethical implications of these models. Our study highlights the intrinsic biases of current models in equitably representing different social identities. Such disparities in accuracy could have adverse effects on equity for how certain groups can use and engage with AI systems.

Moreover, there is a major ethical concern that these models may perpetuate negativity against certain groups, such as Black and Asian individuals, and perhaps especially intersectional groups (e.g., Black and Asian women). Given that the intrinsic biases embedded in the representational spaces of VLMs also lead to differential extrinsic outputs, using VLMs for tasks involving these groups content result in further marginalization and discrimination against these groups. Indeed, when AI users get outputs that reinforce (rather than challenge) their biases, they may come to view those biases as more acceptable and normative \cite{vlasceanu2022propagation}.

In conclusion, our study highlights that intrinsic representations related to social group identity and valence are biased in pretrained VLMs, and that they propagate such biases to downstream zero-shot applications. It is crucial to develop and use models that accurately represent all social identities in a fair and unbiased manner to avoid perpetuating stereotypes and biases against certain groups and to prevent further discrimination and marginalization. An important ethical consideration in this context is the balance between accuracy and fairness. It may be necessary to deliberately interrupt the propagation of biases from intrinsic representational spaces to achieve fairness, even if this intervention compromises some aspects of the model's performance. This trade-off reflects a crucial ethical stance where the value of fairness and inclusivity is prioritized over optimizing for accuracy alone. Such a perspective is essential in developing AI technologies that are not only advanced in their capabilities but are also aligned with broader societal values and ethical principles.

\section{Acknowledgments}
This work was supported by the U.S. National Institute of Standards and Technology (NIST) Grant 60NANB23D194. Any opinions, findings, and conclusions or recommendations expressed in this material are those of the authors and do not necessarily reflect those of NIST.
\bibliography{acl_latex}

\appendix

\appendix
\label{sec:appendix}

\section{Data}
\label{sec:appData} 
We utilize two types of image datasets: one representing social group identity in terms of race and gender, and the other capturing valence (positivity/negativity) as rated by human annotators. We employ valence ratings as a metric to analyze bias due to its nature of generating strong and well understood signals in both image and text \cite{toney-caliskan-2021-valnorm,wolfe2022contrastive}. Additionally, we employ text datasets to complement the image datasets. The sections below detail the specific characteristics of these datasets, the rationale behind their selection, and how they are used following from Section 3 of the Main text. 

\subsection{Chicago Face Database (CFD).} We chose the Chicago Face Database (CFD) \cite{ma2015chicago} for its reliable, controlled source of self-identified human face images, providing clear group signals—unlike FairFace \cite{karkkainen2021fairface}, which contains visually noisy images and lacks self-identified labels, leading to potential bias in annotations. These images are provided with consent of participants to be shared according to agreed upon terms. All personally identifiable information is anonymised \cite{ma2015chicago} before the dataset is released. 

The CFD dataset consists of 597 self-identified images of men and women participants belonging to 4 different racial demographics, namely, Black, White, Latino/a and Asian. \footnote{Because of the known ambiguity in racial perception surrounding Latino/a faces, the current research focused only on Black, White, and Asian racial groups \cite{garcia2013you}}. The photos in the CFD are high-resolution and standardized so that the face always occupies the same portion of the image and they are all placed against a White background. Subjects were photographed with various facial expressions (happy, angry, and terrified) and with a neutral expression while facing directly to the camera. To maintain consistency with \citet{ho2011evidence} and \citet{wolfe2022evidence}, we use only images with neutral facial expressions. In order to obtain a dataset of significant sample size that, we first expand the dataset by creating within-group image morphs \cite{wolfe2022evidence} as described below. The images are then randomly sampled without replacement. 

\subsubsection{Generation of morphed images}
\label{app:morph}

To begin, we first created all possible pairs of images within each demographic group in the CFD dataset. For example, if we consider the Black men demographic, there were approximately 100 images available. We created 50\% morphs of each possible pair of images, which resulted in around 5,000 morphs for each demographic group.

To generate realistic morphed images, we adopt the state-of-the-art StyleGAN2-ADA3 architecture that is pretrained on the high-quality FFHQ dataset. To normalize images, we crop around the facial region, ensuring that the facial features are in positions comparable to those in the pretraining dataset. 

Using StyleGAN2-ADA3, we train the generator on the source and target images of each pair to produce a morph sequence. Given the standardised, high-resolution photos we feed the GAN, we train it for 125 iterations per image using the default hyperparameters of StyleGAN2-ADA3, after which no appreciable effect is shown \cite{wolfe2022evidence}. For each pair, the trained generator creates a source embedding and a target embedding. Subsequently, we use these embeddings to generate the first and second images of each morph sequence.

Finally, we obtain the projected image embedding from CLIP for each morph. This produces a dataset of ~30,000 morphed images with corresponding embeddings, each of which is distinct and diverse. Note that due to copyright issues, we are not able to share generated samples from CFD in the Appendix, but can do so upon request.

\textbf{Operationlisation of CFD} 
We randomly select 1,000 (this number is chosen for its significant sample size as well as computational feasibility) images per intersectional group without replacement, resulting in a total of 6,000 images. In experiments where CFD is used in the group SC-EAT, 140 images per group are selected, including 840 separate images from the SC-EAT group of interest for the two attribute sets. For text-to-image (TTI) retrieval, all 6,000 equally balanced CFD images are used. When CFD is the target dataset for valence SC-EAT and image-to-text (ITT) retrieval, all 6,000 images are utilized.

We want to note that the GAN generative step is used strictly for data augmentation within a group, minimizing the risk of introducing spurious cues about race or gender. While generated images can sometimes introduce artefacts, our approach carefully confines morphs within a single demographic cluster so that any potential artefact is minimal and does not conflate racial and gender categories.

\subsection{Open Affective Standardized Image Set (OASIS) Database.} The Open Affective Standardized Image Set (OASIS) \cite{kurdi2017introducing} comprises 900 color images, depicting a broad spectrum of themes such as humans, animals, objects, and scenes. These images were rated by 822 human participants on valence (degree of positivity or negativity) and arousal (intensity of the valenced response). The 900 images are used in the current experiment to provide non-group-related valence signals from naturalistic image inputs. All images were standardised to 500 X 400 pixels through scaling/cropping processes similar to \citet{wolfe2022evidence}. 

\textbf{Operationlisation of OASIS}
All 900 images are used across experiments where OASIS is required for group SC-EAT or ITT retrieval. In experiments involving valence SC-EAT, we use the top 25 pleasant and top 25 unpleasant OASIS images, sorted by valence ratings, as the two attribute sets. For TTI retrieval, all 900 images are employed. Utilizing the entire lexicon and valence spectrum allows us to provide comprehensive insights into bias propagation, with implications extending to the factuality of retrieval results.

\subsection{Textual Templates and Lexica.} 
Our research requires controlled and balanced text datasets to understand how the psycholinguistic and social group properties of textual content influence biases in VLMs. Specifically, we seek to analyze how words and phrases, with known psycholinguistic ground-truth ratings or specific intersectional identities belonging to one of the 6 groups of interest, can affect the propagation of biases in these models. In this study, we adopt the sentence template approach of \citet{may-etal-2019-measuring} and \citet{tan2019assessing}, utilizing ``semantically bleached templates.'' These templates are designed to be semantically neutral, meaning they do not contribute novel semantic information but ensure that the target word phrases are placed within similar syntactic frames. This approach allows the values associated with the target words to convey the psycho-semantic characteristics of the sentence.

A ``target word'' refers to a specific word inserted into a template, designed to elicit a psycho-semantic response. For example, in the template ``This is the word [WORD],'' the target word replaces ``[WORD]'' and serves as the primary variable in our experiment. In our study we use 6 sentence templates derived from \citet{may-etal-2019-measuring} which are listed in Table \ref{tab:templates-and-words}. 

For experiments involving the use of valence-rated sentences, we employ the NRC-VAD psycholinguistic lexicon \cite{mohammad2018obtaining}, comprising approximately 20,000 English words, with each word rated by 6 human participants on three key dimensions: valence, arousal, and dominance (VAD). This comprehensive lexicon provides us with a robust database of words grounded in reliable, human-rated psycholinguistic data.

For experiments that involve sentences with group signals, we employ a curated list of group labels derived from \citet{charlesworth2022historical}. Here, the target word is a combination of a race-identifying word followed by a gender-identifying word. An example sentence is, ``This is the word Black Woman''. The full list of group words is in Table \ref{tab:templates-and-words}.

\textbf{Operationlisation of NRC-VAD Lexicon Sentences} We use 20,000 words across six sentence templates, with all results averaged across templates. All 20,000 sentences per template are used when the NRC-VAD lexicon is required for group SC-EAT or TTI retrieval. In experiments requiring valence SC-EAT, the top 25 pleasant and top 25 unpleasant sentences, sorted by valence ratings, are used as the two attribute sets. For ITT retrieval, all 20,000 sentences per template are employed. The comprehensive use of the lexicon and valence spectrum provides critical insights into bias propagation and its broader implications.

\textbf{Operationlisation of Group Label Sentences}

We use 5,184 sentences across six sentence templates, with results averaged across templates. When Group Label sentences are the target dataset for valence SC-EAT and TTI retrieval, all 5,184 sentences are used. For group SC-EAT, where Group Label sentences measure group association, 140 sentences per group are randomly selected, including 840 separate sentences from the SC-EAT group of interest for the two attribute sets. For ITT retrieval, all 5,184 sentences are utilized.

\begin{table*}[htbp]
\centering
\caption{Sentence templates and group words used for bias measurement in VLMs. Sentence templates are adopted from \citet{may-etal-2019-measuring}, and group words are selected based on \citet{charlesworth2022historical} These elements form the basis of the text datasets employed in our study to investigate the propagation of social biases.}
\label{tab:templates-and-words}
\begin{tabular}{|l|p{10cm}|}
\hline
\textbf{Type} & \textbf{Content} \\ \hline
Templates & ``This is the word [WORD]'', ``That is the word [WORD]'', ``There is the word [WORD]'', ``Here is the word [WORD]'', ``They are the word [WORD]'', ``Those are the word [WORD]'' \\ \hline
Gender Words & woman, daughter, mother, sister, grandmother, niece, female, girl, madam, aunt, maiden, queen, man, son, father, brother, grandfather, nephew, male, boy, sir, uncle, gentleman, king \\ \hline
Race Words & black, blacks, black-american, afro-caribbean, dark-skinned, jamaican, african, africans, ethiopian, ethiopians, african-american, afro-american, white, whites, british, caucasian, caucasians, light-skinned, american, americans, european, europeans, englishman, englishmen, asian, asians, asian-american, asian-americans, chinese-american, japanese-american, chinese, asiatic, japanese, korean, koreans, korean-american \\ \hline
\end{tabular}
\end{table*}

\section{Models}
\label{sec:Models}
Following from Section 5 of the Main text, the following details the choice of open-source models used in our experiments.

\noindent \textbf{Vision-Language Models -- CLIP (B-32 and L-14).} \citet{radford2021learning} use the WebImageText corpus (WIT), a web scrape made up of 400 million images and related captions to train CLIP. The query list is created by \citet{radford2021learning} by utilising all words that appear at least 100 times in English Wikipedia, as well as word bi-grams with high pointwise mutual information from Wikipedia, the titles of Wikipedia articles, and all WordNet synsets. As part of the creation process, they look for image-text pairs whose texts contain one of a set of 500,000 queries in an effort to cover as many visual notions as feasible. 
The significance of CLIP, particularly its B-32 and the significantly larger L-14 versions, lies in its zero-shot learning capabilities (achieving zero-shot accuracy of 63.2\% and 76.2\%  respectively) on ImageNet \cite{radford2021learning}. These models can process and interpret images and their associated texts without needing task-specific training, demonstrating a level of generalization that is highly applicable in various settings \cite{Briggs_Laura_2022}. 

\noindent \textbf{BLIP2.} BLIP-2 \cite{li2023blip} is a vision-language pre-trained model that integrates off-the-shelf frozen image encoders and LLMs. It includes a Querying Transformer (Q-Former) trained in two stages: vision-language representation learning from a frozen image encoder and vision-to-language generative learning from a frozen LLM. 
It's trained on a dataset comprising 129 million images, including a subset of COCO, Visual Genome, CC3M, CC12M, SBU, and LAION400M datasets. We study BLIP-2 due to its efficiency and high performance in zero-shot tasks. For instance, it achieved state-of-the-art zero shot accuracy of 65\% on VQAv2 \cite{li2023blip}.

\section{Approach}\label{sec:appApproach}
The following section elaborates on the definition and formulation of the SC-EAT taken from \citet{caliskan2017semantics} and \citet{steed2021image}.

\section{Details of Experiments}
\label{sec:appExp}
This section supplements the results mentioned in Section 5 in the Main text and contains the details for the reproducibility of all experiments, including specific datasets and parameter settings used. Figure \ref{fig:doe} contains the reproduction details of each of the 8 bias and association propagation experiments.

\subsection{Initial Assessment of Intrinsic Biases in Vision-Language Models} 
\label{app:initial_res}
\begin{figure*}[ht!]
      \begin{minipage}[c]{0.46\textwidth}
\centering  
\includegraphics[width=0.8\linewidth]{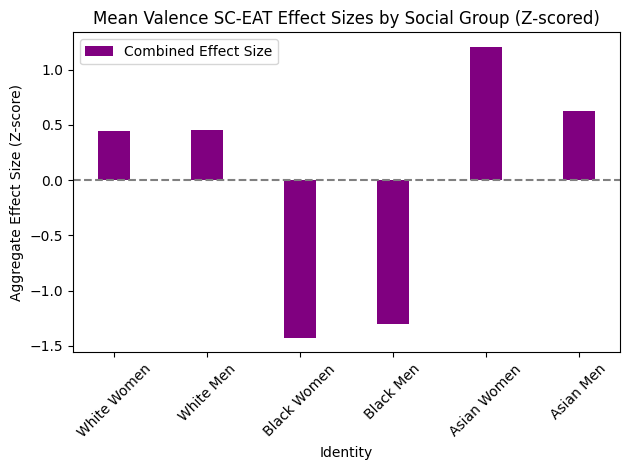}
\centering
  \caption{Valence based Intrinsic Bias Visualization through SC-EAT Effect Sizes: This figure presents the normalised mean of SC-EAT effect sizes for different social groups obtained in experiments 1-a and 1-b. A negative effect size indicates a stronger association of the group with negative valence than positive. The data reveals intrinsic biases within the models, where notably, Black Women are depicted with the highest negative association, as indicated by the most negative effect size. In contrast, Asian Men and Women exhibit the least negative associations. This graphically demonstrates the prevalence of intrinsic bias in the representational spaces of VLMs, with varying degrees of negativity associated with each group.}
  \label{fig:sc-eat-effect-sizes}
       \end{minipage}
\begin{minipage}[c]{0.49\textwidth}
\centering
\includegraphics[width=\linewidth]{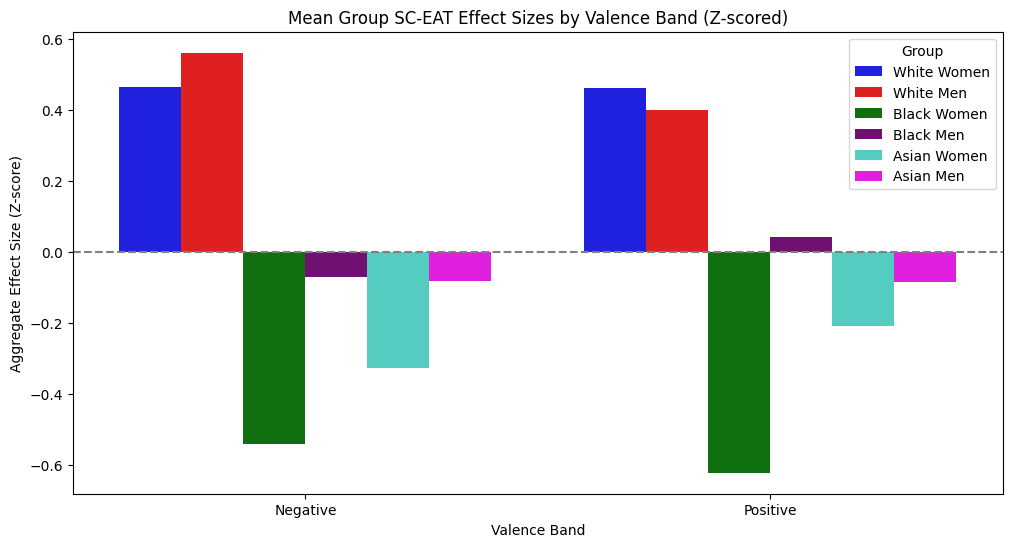}
  \caption{Group-based Intrinsic Bias Visualization through SC-EAT Effect Sizes: This figure presents the normalised mean of SC-EAT effect sizes for high/low valence bands obtained in experiments 2-a and 2-b. We categorise images/text with valence $\geq$ 0.5 as positive and negative otherwise, on a scale of 0 to 1. A positive effect size indicates a stronger association of the valenced item with the indicated group as opposed to any group at random. The data reveals intrinsic biases within the models, where notably, White Men and Women are depicted with the highest positive and negative associations. In contrast, Black Women exhibit the least magnitude in associations to valenced images/text. This graphically demonstrates the prevalence of group-based intrinsic bias in the representational spaces of VLMs. }
  \label{fig:sc-eat-effect-sizes-2}
  \end{minipage}
  \hspace{8mm}
\end{figure*}

\begin{figure*}[ht]
  \centering
  \includegraphics[width=\linewidth]{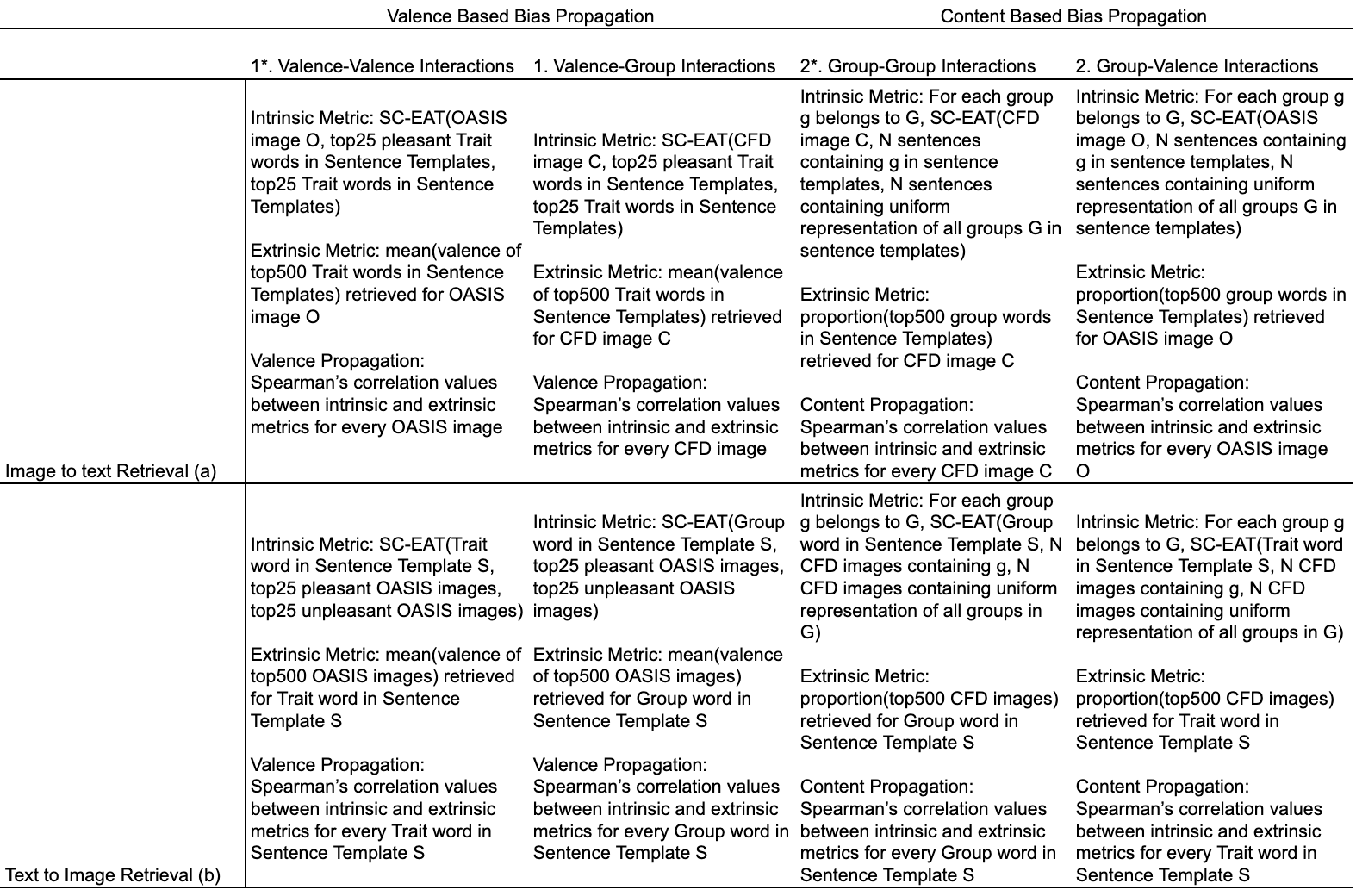}
  \caption{Figure details the specifics of the 8 experiments used to show the propagation of intrinsic bias to extrinsic zero-shot retrieval tasks.  The experiments share a common framework for measuring intrinsic bias (e.g., the effect size of bias that captures the differential association of a social group image with negatively valenced sentences over positively valenced sentences), followed by an assessment of extrinsic bias (e.g., the likelihood of retrieving negatively valenced text for a given group image prompt), and then correlating these two metrics. The experiments vary in their focus on the direction of bias propagation (four each on image-to-text and text-to-image) and the type of content analyzed (four on valence or rating of positivity/negativity, and four on group signals)}
\label{fig:doe}
\end{figure*}

Figures \ref{fig:sc-eat-effect-sizes} and \ref{fig:sc-eat-effect-sizes-2} highlight significant intrinsic biases within VLMs. Figure 8 presents the aggregate valence-based SC-EAT effect sizes for different social groups obtained in experiments 1-a and 1-b. Figure 9 presents the aggregate group-based SC-EAT effect sizes for high (valence $\geq$ 0.5) and low (valence $<$ 0.5) valence bands across experiments 2-a and 2-b.

\subsection{Propagation of Bias Through Valence and Group Signals}
\noindent \textbf{Baseline Propagation of Intrinsic Valence Associations to Extrinsic Valence Outcomes.}

This experiment serves as a baseline to measure how isolated valence signals in text and valence signals in images are propagated through VLMs. We follow previous works such as VAST \cite{wolfe2022vast} and ValNorm \cite{toney-caliskan-2021-valnorm} that use intrinsic valence signals to evaluate the intrinsic quality of word embeddings and language models. First, for the image-to-text direction (1*-a), we measure the correlation between the intrinsic valence of the image and the mean valence of the top-500 sentences retrieved based on that image. For instance, an image with a high intrinsic positive valence (e.g., roses, butterflies, etc) should ideally retrieve text with similarly positive valence (e.g., ``This is the word pretty''). The intrinsic measure here is the SC-EAT score of each of the 50 most valenced OASIS images (top 25 positive and top 25 negative); the extrinsic measure here is the average valence rating (from human valence ratings) of the top 500 sentences associated with the specific input OASIS image.

Second, the text-to-image (1*-b) direction mirrors 1*-a and explores how intrinsic valence signals in text correlate with the aggregate valence of retrieved images based on the text input. Specifically, the intrinsic measure is the SC-EAT score of the 50 most valenced words in sentences (top 25 positive and top 25 negative words, as described in Section \ref{sec:appData}); the extrinsic measure here is the average valence rating (from human valence ratings) of the top 500 OASIS images associated with the input sentence.

\noindent \textbf{Baseline Propagation of Intrinsic Group Associations to Extrinsic Group Outcomes.}
This set of experiments seeks to establish a baseline for understanding the simple propagation of group content. In the image-to-text setting (2*-a), we test the correlation between the representation of a specific social group in an image and the group representation in the retrieved text. The intrinsic measure here is the image-text group SC-EAT where we find the association between a given image (from the CFD) and sentences including a single intersectional group (versus sentences that represent all intersectional groups). This method essentially parallels a one-vs-all machine learning verification task, where the model's ability to differentiate one group identity from all others at the representational level is evaluated. Extrinsically, for the same CFD image, we obtain the proportion of group-identified sentences retrieved that are ``correct'' (i.e., the same intersectional identity as the text prompt).  For example, if an input image is of a ``Black Woman,'' then extrinsically we assess how often the text retrieved accurately reflects Black women concepts.

In the text-to-image direction (2*-b) we test whether text representing a social group retrieves images with that same group representation from the CFD. The intrinsic measure is the text-image SC-EAT between a given text prompt with a specific group signal and images representing the single intersectional group identity (versus images containing all intersectional groups uniformly). Extrinsically, we then record the proportion of images that ``correctly'' represent the specific social group retrieved for text prompt. For clarity, ``correct'' in this context means that the retrieved images accurately mirror the specific social group identity mentioned in the text. For instance, if the text prompt is about ``Asian Men,'' the extrinsic measure evaluates how often the retrieved images indeed depict Asian men. 

As in all other studies, we quantify content/group propagation using Spearman's correlation values between the intrinsic SC-EAT group based associations of the image/text with the extrinsic proportion of sentences/images correctly retrieved for each group.

\noindent \textbf{Propagation of Intrinsic Group-based Valence Associations to Extrinsic Valence Outcomes.}
This experiment is our primary bias propagation analysis as it focuses on how biased associations between group signals (in text or image) and valenced attributes are propagated through VLMs. For the image-to-text direction (1-a), we consider how intrinsic group signals in images influence the valence of retrieved text. The image-text SC-EAT measures the intrinsic valence effect size of each image belonging to a group identity (from the CFD). Then, extrinsically, for the same image, we calculate the mean valence of the top-500 sentences retrieved, averaged over the templates. 

In the text-to-image direction (1-b) we examine how text associated with social groups influences the valence of retrieved images. Intrinsically, this is measured through the text-image SC-EAT by computing the effect size of valenced associations for each sentence containing an intersectional group identity with valence-rated images (from OASIS). Then, extrinsically, for the same sentence, we calculate the mean valence of the top-500 valence-rated OASIS images retrieved.

\noindent \textbf{Propagation of Intrinsic Valence-based Group Associations to Extrinsic Group Outcomes.}
Finally, and in parallel to our primary analysis of group-to-valence propagation, we consider valence-to-group propagation or, more specifically, how intrinsic valence associations of images/text as measured by SC-EAT lead to differential retrieval of group signals. In the image-to-text direction (2-a), we investigate whether stronger intrinsic valence ratings in images correlate with a higher likelihood of retrieving text related to marginalized groups. Intrinsically, we use the image-text group SC-EAT to find the association of a given valenced image (from OASIS) with text containing group terms belonging to a single intersectional identity (versus text uniformly representing all intersectional groups). Extrinsically, we obtain the corresponding proportion of group-identified sentences that are retrieved for the same OASIS image for each of the 6 intersectional social groups. 

In the opposite, text-to-image direction (2-b) we test if text with strong valence ratings retrieves images predominantly featuring certain social groups. The intrinsic measures are the SC-EAT associations between a given valenced text prompt and images from a specific social group (versus images of all groups uniformly that were selected through random balanced sampling without replacement). Extrinsically, we calculate the proportion of images that represent a social group retrieved for the same single intersectional group text prompt.

\end{document}